\documentclass[conference]{IEEEtran}
\IEEEoverridecommandlockouts
\usepackage{times}

 \usepackage{makecell}
\usepackage{graphicx}
\usepackage{amsmath}
\usepackage{algorithm}
\usepackage{algorithmic}
\usepackage{graphicx}
\usepackage{subfigure}

\usepackage{times}
\usepackage{epsfig}
\usepackage{graphicx}
\usepackage{amsmath}
\usepackage{amssymb}

\usepackage{bm}
\usepackage{graphicx}
\usepackage{times}

\usepackage{amsmath}
\usepackage[english]{babel}
\usepackage{graphicx}
\usepackage{epsfig}
\usepackage{epstopdf}
\usepackage{amsmath}
\usepackage[english]{babel}
\usepackage{graphicx}
\usepackage{graphicx}
\usepackage{epsfig}
\usepackage{epstopdf}
\usepackage{amssymb}
\usepackage{subfigure}
\usepackage{color}
\def\BibTeX{{\rm B\kern-.05em{\sc i\kern-.025em b}\kern-.08em
    T\kern-.1667em\lower.7ex\hbox{E}\kern-.125emX}}
\begin{document}

\title{Using  Gradient based multikernel Gaussian Process and Meta-acquisition function   to Accelerate SMBO}
 
\author{\IEEEauthorblockN{Daning Cheng \IEEEauthorrefmark{1}\IEEEauthorrefmark{2},
Hanping Zhang\IEEEauthorrefmark{3},
Fen Xia\IEEEauthorrefmark{3}, 
Shigang Li\IEEEauthorrefmark{1} and
Yunquan Zhang\IEEEauthorrefmark{1}}
\IEEEauthorblockA{\IEEEauthorrefmark{1}SKL of Computer Architecture, Institute of Computing Technology, CAS, China\\
Email:  \{chengdaning, lishigang, zyq\}@ict.ac.cn}
\IEEEauthorblockA{\IEEEauthorrefmark{2}University of Chinese Academy of Sciences
}
\IEEEauthorblockA{\IEEEauthorrefmark{3}WiseUranium corp.\\
{\{ Xiafen, Zhanghanping \}@ebrain.ai}}
 }

\maketitle

\begin{abstract}
	Automatic machine learning (automl) is a crucial technology in machine learning.  Sequential model-based optimisation algorithms (SMBO) (e.g., SMAC, TPE)  are state-of-the-art hyperparameter optimisation methods in automl. 
	
	However, SMBO does not consider known information, like the best hyperparameters high possibility range and gradients. In this paper, we accelerate the traditional SMBO method and name our method as accSMBO. In accSMBO, we build a gradient-based multikernel Gaussian process with a good generalisation ability  and we design meta-acquisition function which encourages that SMBO puts more attention on the best hyperparameters high possibility range.  In L2 norm regularised logistic loss function experiments, our method exhibited state-of-the-art performance.
\end{abstract}


\section{Introduction}
Automatic machine learning (automl) is a key technology in machine learning:  Current machine learning models require enormous numbers of hyperparameters. Thus, hyperparameter optimisation is the key part of automl. Sequential model-based optimisation (SMBO) is a state-of-the-art hyperparameter optimisation algorithm frame. However, traditional SMBO does not consider hyperparameter known information. Intuitively known information can accelerate the convergence process. In this paper, we use (1) hyperparameter gradient (2) regular patterns of hyperparameters' performance  to accelerate SMBO.  

In this paper, we propose a novel algorithm: accSMBO. AccSMBO uses the following two methods to optimise SMBO: (1). We use a gradient-based multikernel Gaussian process regression to fit the observed values and the observed gradient values. Consequently, SMBO builds the response surface of the hyperparameter-performance curve faster with a good generalisation ability and less computational burden. (2). We propose meta-acquisition function: We build an empirical probability density function, abbr. EPDF, based on the metalearning dataset. At each SMBO iteration, we adjust the candidate hyperparameter towards the ranges with a high probability of the best hyperparameters based on EPDF. The above methods accelerate SMBO and achieve satisfactory results. In the experiments, accSMBO improves the convergence speed by 140\% to 300\% in  epoch compared to the SMAC algorithm on different datasets.

Our main contributions are summarised as follows. (1) We propose a novelty gradient-based multikernel Gaussian process regression to accelerate the SMBO.  (2)  We propose the meta-acquisition function which encourages SMBO to explore hyperparameters on best hyperparameters high possibility range.  (3) In L2 norm experiments, our method achieves convergence 140\% to 300\% faster in  epoch than SMAC on datasets of different scales. It outperformed the previous best hyperparameter optimisation approach.

\section{Related works}

Several automatic machine learning tools have been developed, such as Google's AutoML, autoWEKA\cite{thornton2013auto-weka:} and autosklearn in recent years. Those tools contain various types of hyperparameter optimisation algorithms such as probabilistic methods (e.g., Bayesian optimization\cite{thornton2013auto-weka:}\cite{Hutter2012Sequential}  \cite{Jones1998Efficient} \cite{Regis2005Constrained}), random optimization methods (e.g., grid search, heuristic algorithms, and Neural Networks\cite{mendoza2016towards}), Fourier analysis (e.g., Harmonica\cite{Hazan2017Hyperparameter}), and decision-theoretic methods (e.g., the Successive Halving (SH) algorithm and  Hyperband\cite{Li2016Hyperband}). 

SMBO algorithms\cite{Hutter2012Sequential}, are currently the most widely used Bayesian optimization in automl \cite{thornton2013auto-weka:} \cite{feurer2015efficient}\cite{Hutter2012Sequential}. SMAC\cite{Hutter2012Sequential},  TPE  \cite{Bergstra2011Algorithms} and Gaussian-process-based SMBO \cite{Snoek2012Practical} are  the state-of-the-art SMBO algorithms. In automl, metalearning\cite{Brazdil2009Metalearning} and gradient-based hyperparameter optimization\cite{Pedregosa2016Hyperparameter}\cite{Do2007Efficient} \cite{Wu2017Bayesian}  are the hotly debated topics.

\textbf{Problem}  Current SMBO algorithms do not make full use of known information in their iteration processes. 

\section{Background}
\subsection{Problem Setting}
Hyperparameter optimization focuses on the problem of learning a performance function $f: \mathcal{X} \mapsto \mathcal{Y}$ with a finite $\mathcal{Y}$. A $learning$ $algorithm$ $\mathcal{A}$ exposes $hyperparameters$ $\lambda \in \Lambda$  that change the way the learning algorithm $A(\lambda)$ operates.

For a given learning algorithm $\mathcal{A}$ and a limited amount of training data $\mathcal{D} = \{ (x_1,h_1),(x_2,y_2),...,(x_m,y_m) \}$, the goal of hyperparameter optimization is to minimize the performance function $f$, which is estimated by splitting $\mathcal{D}$ into disjoint training and validation sets, 
$\mathcal{D}_{train}^{(i)}$ and $\mathcal{D}_{vail}^{(i)}$, respectively. The performance function $f$ is applied by $A$ with  $\lambda^* \in \Lambda$ to $\mathcal{D}_{train}^{(i)}$; then, the predictive performance of these functions on $\mathcal{D}_{vail}^{(i)}$ is evaluated. This approach allows the hyperparameter optimization problem to be written as follows:

\begin{align}
	&\lambda^* \in \mathop{argmin}_{\lambda \in \Lambda} f(\lambda) \triangleq g(A(\lambda),\lambda) \notag \\
	&=\frac{1}{k} \sum_{1}^{k} \mathcal{L}(A(\lambda),\mathcal{D}_{train}^{(i)},\mathcal{D}_{vail}^{(i)}) \notag \\
	&s.t. A(\lambda) \in \mathop{argmin}_{\textbf{model} \in \mathbb{R}} h(\lambda,\textbf{model}) \label{define}
\end{align}

where $ \mathcal{L}(A,\mathcal{D}_{train}^{(i)},\mathcal{D}_{vail}^{(i)})$ is the loss achieved by $A$ when trained on $\mathcal{D}_{train}^{(i)}$ and evaluated on $\mathcal{D}_{vail}^{(i)}$. We use k-fold cross-validation [18], which splits the
training data into $k$ equal-sized partitions $\mathcal{D}_{vail}^{(i)}$

\subsection{Basic Information}
\textbf{SMBO algorithm frame and its state-of-art algorithm} The SMBO is a black-box optimisation algorithm frame. The SMBO algorithm frame, algorithm \ref{SMBO}, does not possess complete information concerning $f(\cdot)$. It requires only sample $(\lambda, f(\lambda))$ values to build $f(\lambda)$'s model $\mathcal{M}_L$ ( In some works,  $\mathcal{M}_L$ is named as response surface\cite{Jones1998Efficient}\cite{Regis2005Constrained}). 
\begin{algorithm}[!tb]
	\caption{SMBO}
	\label{SMBO}
	\begin{algorithmic}
		\STATE {\bfseries Input:} {hyperparameter history $\mathcal{H}$, initial hyperparameter $\lambda$}
		\STATE {\bfseries Output:} {$\lambda$ from $\mathcal{H}$ with minimal $c$} 
		
		\STATE 
		\STATE step 0: Choose initial value $\lambda_0$ and its $f(\lambda_0)$ and add $(\lambda_0,f(\lambda_0))$ into $\mathcal{H}$
		\REPEAT
		\STATE step 1: Update $\mathcal{M}_L$ given $\mathcal{H}$ and compute   $acquisition$ $function$.  
		\STATE step 2:  Gain the hyperparameter candidates from  $acquisition$ $function$.
		\STATE step 3: $\lambda\leftarrow$ select the best candidate hyperparameter in candidates from step 2 \;
		\STATE step 4: Compute $ f(\lambda)$\;
		\STATE step 5: $\mathcal{H}\leftarrow\mathcal{H} \bigcup \{ (\lambda,f(\lambda))\}\} $\;    
		\UNTIL{the time budget has not been exhausted}
	\end{algorithmic}
\end{algorithm}

\begin{table}
	\caption{State of the art algorithm for SMBO algorithm frame}
	\label{SMBO_fill}
	\begin{tabular}{|c|>{\makecell*[c]}c|}
		\hline
		Algorithm & Step in SMBO \\
		\hline
		\makecell[tl]{MetaLearning \cite{Vanschoren2018Meta} \cite{Brazdil2009Metalearning} \\ Random selection} & Step 0 \\
		\hline
		\makecell[tl]{
			\textcolor{red}{Random Forest \& Gaussian Process(SMAC)}\cite{Hutter2012Sequential} \\
			Tree-structured Parzen Estimator(TPE)\cite{Bergstra2011Algorithms}\\
			Neural Network\cite{mendoza2016towards}\\
			Fourier analysis\cite{Hazan2017Hyperparameter}\\
			Gaussian Process \cite{Snoek2012Practical}\\ 
		} & Step 1: $\mathcal{M}_L$ \\
		\hline
		\makecell[tl]{Probability of Improvement \cite{kushner1964a}\\
			\textcolor{red}{Expected Improvement}\cite{Jones1998Efficient} \\ 
			d-KG (contains gradient information) \cite{Wu2017Bayesian}\\
			GP Upper Confidence Bound\cite{Srinivas2009Gaussian}} 
		& \makecell[tl]{Step 1: \\ $acquisition$ $function$} \\
		\hline
		\makecell[tl]{hyperband\cite{Li2016Hyperband} \\ \textcolor{red}{Intensify process}\cite{Hutter2012Sequential}} & Step 3 \\
		\hline
	\end{tabular}
\end{table}

The core of SMBO is building a model $\mathcal{M}_L$ that captures the dependence on the loss function $\mathcal{L}$ for the various hyperparameter settings and using $acquisition function$ to choose candidate hyperparameters. To make our paper clear, we use  "AC function ($ac(\lambda)$)" as the abbr. of $acquisition$ $function$. 

Researchers proposed different algorithms to fill the SMBO algorithm frame. Most of these state of the art algorithms are presented at table \ref{SMBO_fill} and the following descriptions.

To make $\mathcal{H}$ has a good initial value, researchers proposed metalearning technology. Based on metalearning, it is possible to get the optimal or sub-optimal hyperparameter values at the first epoch. However, to make metalearning technology outperform, it is necessary to make the metalearning dataset large. A large metalearning dataset usually contains the best hyperparameter values for more cases. 

To make $\mathcal{M}$ reflect sampled points trend better, researchers proposed different response surface models. However, those models do not consider gradient or be sensitive to gradient information. In those models, random forest is the most widely used and state of the art methods\cite{ feurer2015efficient}. The SMBO which uses random forest is SMAC. Thus, SMAC is the main benchmark in this paper.

State of the art $acquisition$ $functions$ choices have expected improvement function\cite{Jones1998Efficient} and d-KG function\cite{Wu2017Bayesian}. D-KG  is the $acquisition$ $function$ which uses gradient information. The abbreviation of  expected improvement function is EI function in this paper. Because our method does not care about the choice of $acquisition$ $functions$, we choose the most widely used EI function as our  $acquisition$ $functions$.

The selection of candidates is the main purpose for step 3 in algorithm \ref{SMBO}. The best algorithms for this step are Hyperband\cite{Li2016Hyperband} and the intensify process\cite{Hutter2012Sequential}.  Because our method does not care about how to choose the best candidate, we use the most widely used intensify process in step 3 in our benchmarks.

\textbf{Epoch}
In SMBO, we name one iteration/loop, as one epoch.  One epoch is the smallest unit to measure the performance of SMBO, for SMBO cost almost the same time at each iteration. 

For different code implementation of algorithm and running platform, the time cost is different for an epoch. Therefore, it is better to use the number of the epoch as an index instead of the time when we conduct experiments.

\textbf{Gaussian process ($\mathcal{GP}$)}  
The GP \cite{Rasmussen2010Gaussian} is defined by the property that any finite set of m points $\{ (x_n,y_n) \in \mathcal{X},\mathcal{Y} \}^m_{n=1}$ induces a multivariate Gaussian distribution on $\mathbb{R}^m$. The $n$th point is taken as the function value $f(x_n)$, and the elegant marginalization properties of the Gaussian distribution allow us to compute marginals and conditionals in closed form. The support and properties of the resulting distribution, $\mathcal{N}(m(x),var(x))$, on functions are determined by a mean function $m(x) : \mathcal{X} \mapsto R$ and a positive definite covariance function $k(x,x):\mathcal{X} * \mathcal{X} \mapsto \mathbb{R}$.

\textbf{$\mathcal{GP}$ regression}     $\mathcal{GP}$ based SMBO is the most commonly used SMBO algorithm.  In $\mathcal{GP}$ based SMBO, the $\mathcal{M}_L$ in algorithm \ref{SMBO} is the $\mathcal{GP}$ regression   from $\mathcal{H}$. The original $\mathcal{H}$ is defined as follows:

\begin{equation*}
	\mathcal{H} = \{  (\lambda_1,f(\lambda_1)),...,(\lambda_n,f(\lambda_n),)
	\} 
\end{equation*}

We set $\lambda$ as a vector with $d$ dimensions. $k(x,x')$ is the  kernel (covariance) function. We define the vector $\textbf{f}=(f(\lambda_1), f(\lambda_2),...,f(\lambda_n))^T$, whose dimension is (1*n). We also define the matrix $\bm{\lambda} = ( \lambda_1, \lambda_2,...,\lambda_n)^T$, whose dimension is (d*n), and the kernel matrix $K(\bm{\lambda},\bm{\lambda})$, whose dimension is $(n*n)$, as follows:

\begin{equation*}
	K(\bm{\lambda},\bm{\lambda}) =
	\left[
	\begin{matrix}
		k(\lambda_1,\lambda_1)      &k(\lambda_1,\lambda_2)       & \cdots & k(\lambda_1,\lambda_n)       \\
		k(\lambda_2,\lambda_1)      & k(\lambda_2,\lambda_2)       & \cdots & k(\lambda_2,\lambda_n)       \\
		\vdots & \vdots & \ddots & \vdots \\
		k(\lambda_n,\lambda_1)       & k(\lambda_n,\lambda_2)       & \cdots & k(\lambda_n,\lambda_n)       \\
	\end{matrix}
	\right]
\end{equation*}

The vector $K(\lambda^*,\bm{\lambda})$ is defined as $K(\lambda^*,\bm{\lambda})=(k(\lambda^*,\lambda_1),k(\lambda^*,\lambda_2),...,k(\lambda^*,\lambda_n))$.

In the traditional $\mathcal{GP}$ regression with noise-free observations, the mean $m(\lambda^*) = \sum_{i =1}^{n} \gamma_i k(\lambda_i,\lambda^*)$, i.e.,  $m(\lambda*) = \bm{\gamma}K(\lambda*,\bm{\lambda})$. $\bm{\gamma}= K(\bm{\lambda},\bm{\lambda})^{-1}\bm{f}$. $var(\lambda^*) = k(\lambda^*,\lambda^*) - K(\bm{\lambda},\lambda^*)K(\bm{\lambda},\bm{\lambda})^{-1}K(\bm{\lambda},\lambda^*)$. The $\mathcal{M}_L$ for  $\mathcal{GP}$ based   SMBO is the  $\mathcal{GP}$ which is $\mathcal{N}(m(\lambda*),var(\lambda^*))$.

\textbf{Hyperparameter gradients} 
Many works \cite{Do2007Efficient}\cite{Pedregosa2016Hyperparameter} offer gradient-based hyperparameter optimization methods. The gradient of a hyperparameter can be  calculated as follows\cite{Pedregosa2016Hyperparameter}:

\begin{align*}
	\nabla f&= \nabla_2g+(\nabla A)^T\nabla_1g\\
	&=\nabla_2g-(\nabla^2_{1,2}h)^T(\nabla^2_1h)^{-1}\nabla_1g
\end{align*}

Those approaches use gradient descent methods, which are different from Bayesian optimisation methods.

\textbf{Meta-Learning dataset}
Metalearning is the key technology to accelerate SMBO by offer SMBO an experiential best initial value. Researchers classify those best initial value by the feature of task, objective function and dataset. Metalearning datasets record those best initial values.

Although Metalearning gains great success in hyperparameter optimisation, for most of the cases,  metalearning technology only accelerates SMBO under the condition that the metalearning dataset contains the corresponding hyperparameter values. When the metalearning dataset is not complete, the improvement of metalearning is limited.

\section{Main idea}
\begin{figure*} 
	\caption{Some of those performance functions are close to unimodal functions or monotonic functions }
	\centering
	\subfigure[The relationship between the decision tree depth and the log loss on the rcv1 dataset]{
		\includegraphics[width=0.3\textwidth]{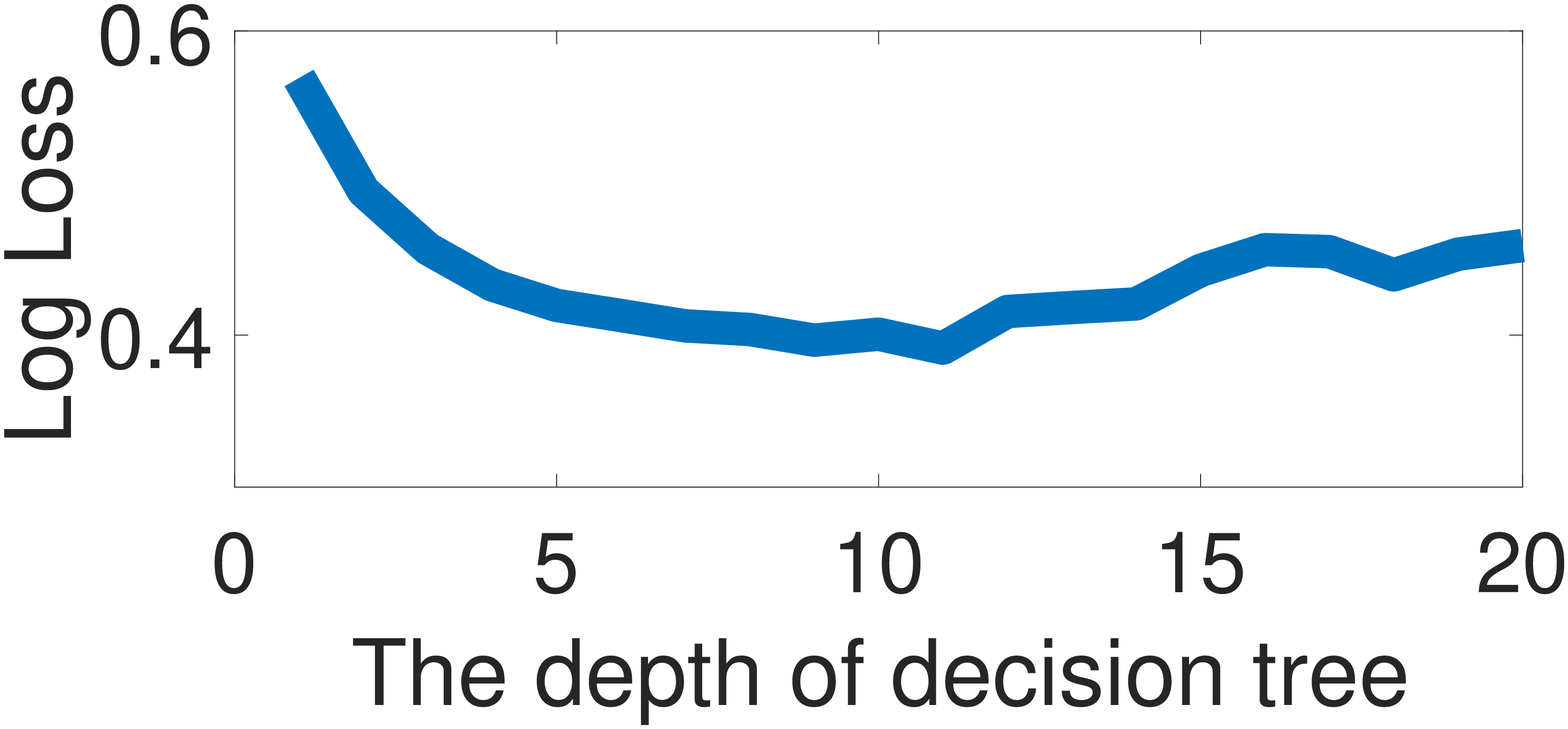}  
	}
	\subfigure[The relationship between the GBDT tree depth and the log loss on the rcv1 dataset]{
		\includegraphics[width=0.3\textwidth]{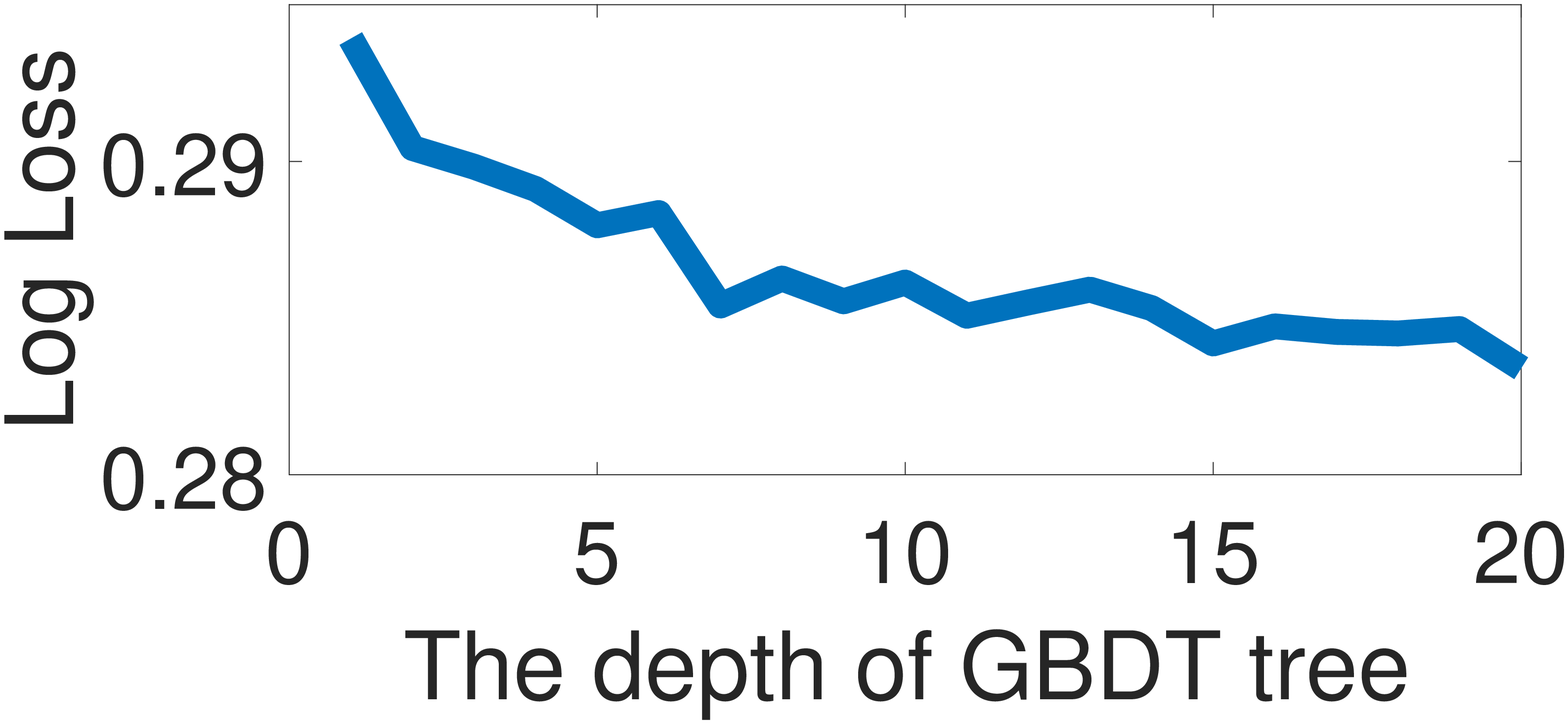} 
	}
	\label{character}
\end{figure*}

\begin{figure*} 
	\caption{The frequency  histogram and its fitting  empirical probability density function of hyperparameter value for f1 norm multi-class task on sparse dataset. The information is collected from the meta-learning dataset}
	\centering
	\subfigure[random\_forest:max\_features]{
		\begin{minipage}[b]{0.3\textwidth}
			\includegraphics[width=1\textwidth]{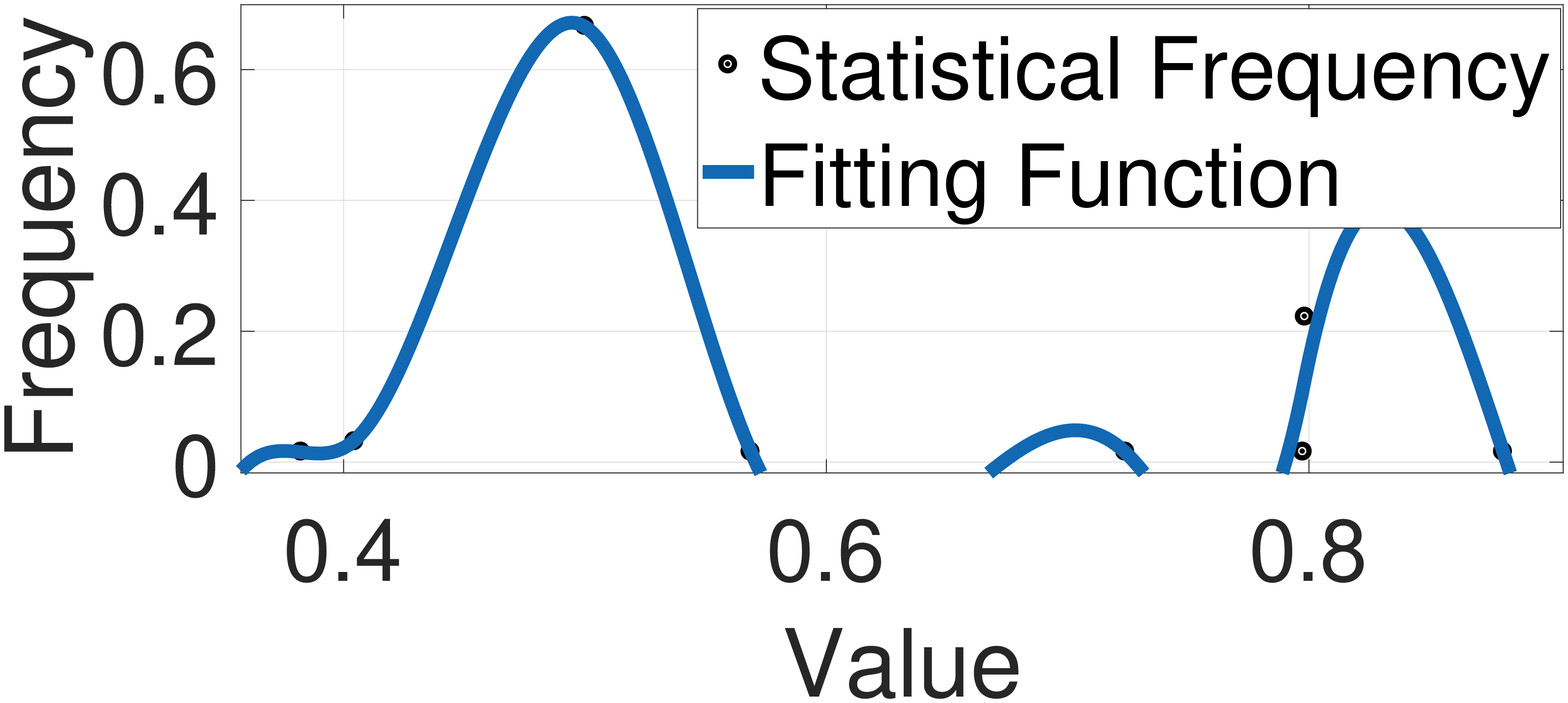}  
			
		\end{minipage}
	}
	\subfigure[random\_forest:min\_samples]{
		\begin{minipage}[b]{0.3\textwidth}
			\includegraphics[width=1\textwidth]{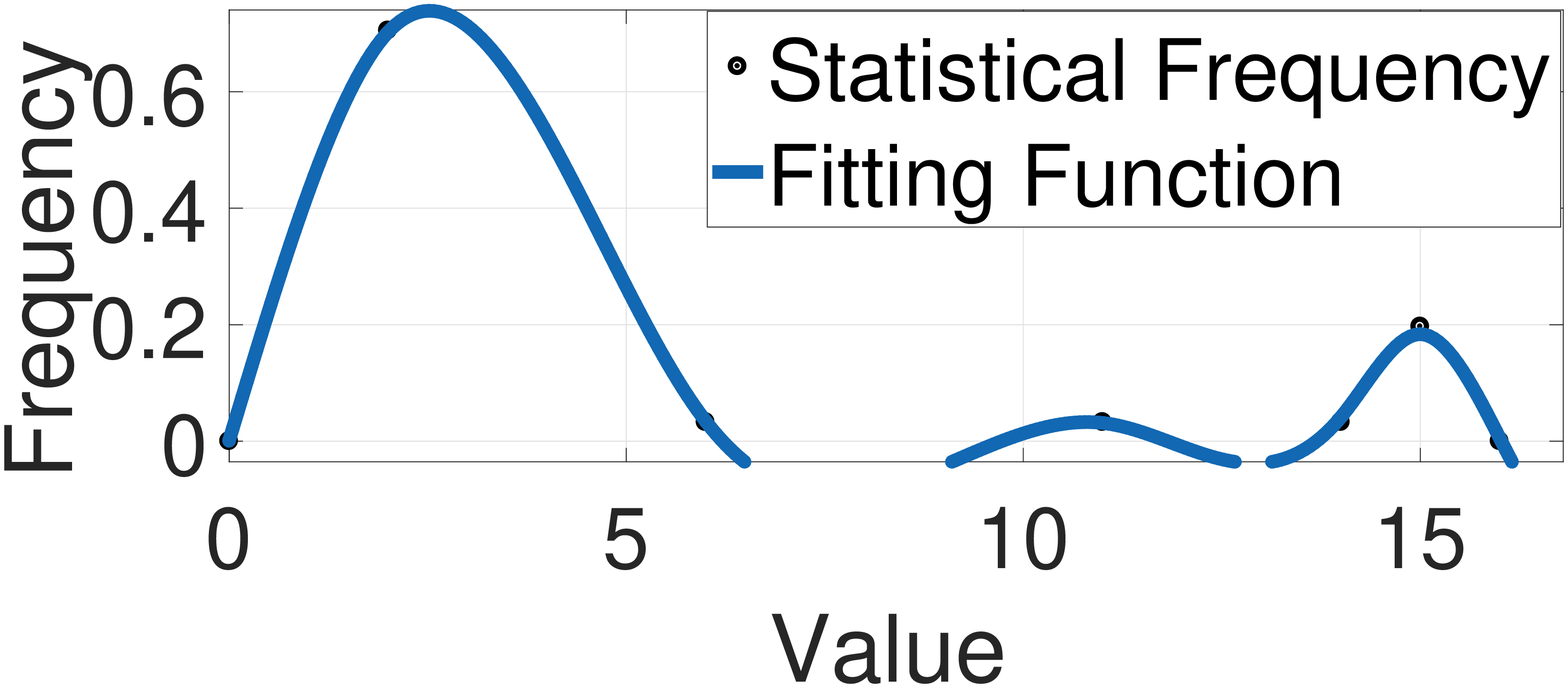} 
			
		\end{minipage}
	}
	\subfigure[liblinear\_svc\_preprocessor:tol]{
		\begin{minipage}[b]{0.3\textwidth}
			\includegraphics[width=1\textwidth]{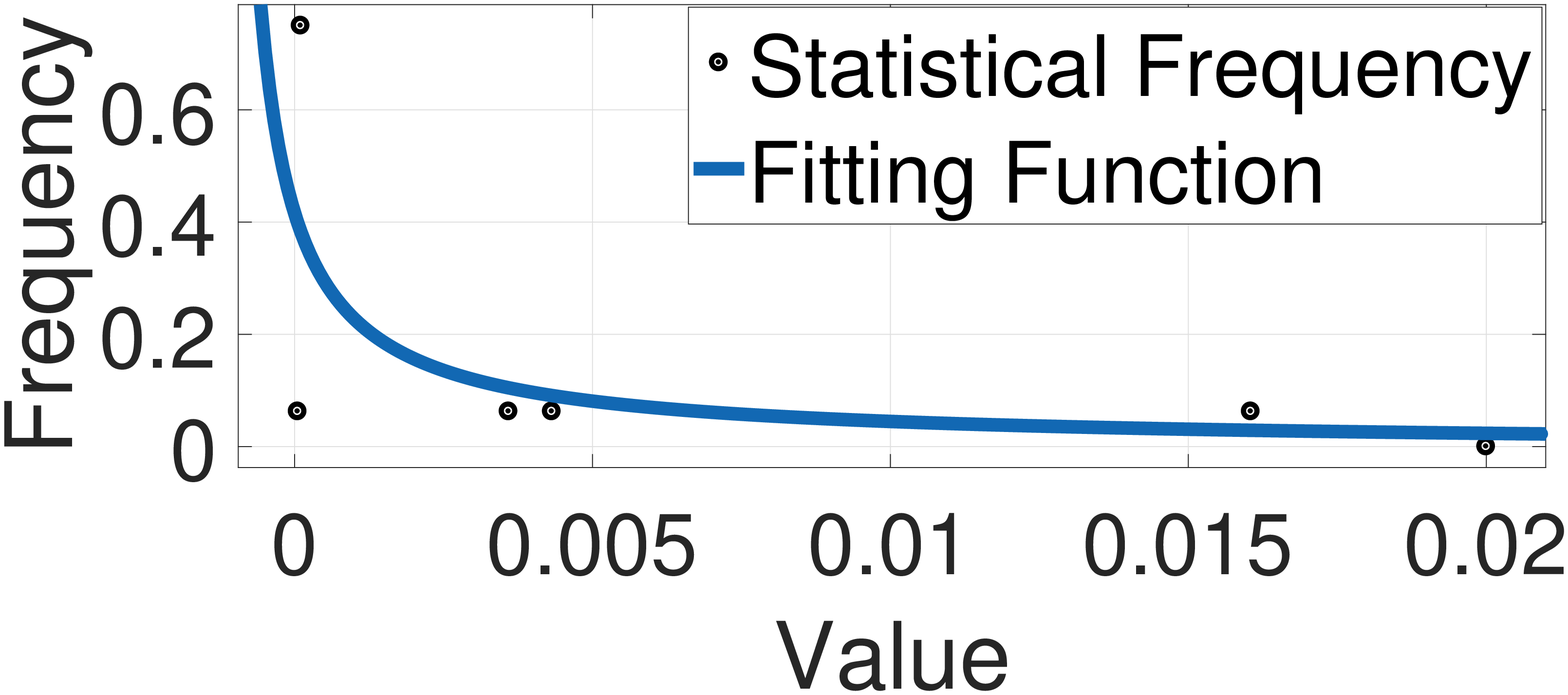} 
			
		\end{minipage}
	}
	\label{character_meta}
\end{figure*}

\begin{algorithm} 
	\caption{AccSMBO}
	\label{accSMBO}
	\begin{algorithmic}
		\STATE {\bfseries Input:} {hyperparameter history $\mathcal{H}$, initial hyperparameter $\lambda$}
		\STATE {\bfseries Output:} {$\lambda$ from $\mathcal{H}$ with minimal $c$} 
		
		\STATE 
		\STATE step 0: Choose initial value $\lambda_0$ and its $f(\lambda_0)$ and add $(\lambda_0,f(\lambda_0))$ into $\mathcal{H}$
		\REPEAT
		
		\STATE step 1: Update $\mathcal{M}_L$ given $\mathcal{H}$ and compute   $acquisition$ $function$.
		\STATE step 2: Using metalearning dataset adjust  $acquisition$ $function$ to $meta$ - $acquisition$ $function$.          
		\STATE step 3:  Gain the hyperparameter candidates from  $meta$ - $acquisition$ $function$.
		\STATE step 4: $\lambda\leftarrow$ select the best candidate hyperparameter in candidates from step 2 \;
		\STATE step 5: Compute $ f(\lambda)$\;
		\STATE step 6: $\mathcal{H}\leftarrow\mathcal{H} \bigcup \{ (\lambda,f(\lambda))\}\} $\;    
		
		\UNTIL{the time budget has not been exhausted}
	\end{algorithmic}
\end{algorithm}
We noticed that for most hyperparameters, the performances of the hyperparameters present the following regular patterns: 1). Generally, the performance of the hyperparameters, like the logloss-hyperparameters curve, is simple, such as a monotonic function or unimodal function. However, locally, those performances are unstable, i.e. full of waves, as shown in Figure \ref{character}. 2). The distribution of the best hyperparameter is not a
uniform distribution. Thus we know some prior, reasonable ranges and the best hyperparameter  is in this range with high probability. It is obviously that metalearning datasets can reflect this information. Figure \ref{character_meta} shows that for the F1 norm multi-class task, the frequency  histogram and its fitting  empirical probability density function for hyperparameter max\_feature and min\_samples in random forest and tol value in svc process.

To accelerate SMBO and make full use of the above regular patterns,  we propose an accelerated SMBO, named as accSMBO, which is illustrated in algorithm \ref{accSMBO}. 

Compared with traditional SMBO, AccSMBO algorithm modifies two parts: 1.) $\mathcal{M}_L$. AccSMBO asks $\mathcal{M}_L$ to reflect the approximate gradients. AccSMBO  builds $\mathcal{M}_L$ which reflects general performance trends with fast speed and 2.) The structure of SMBO.  AccSMBO uses the metalearning dataset in the iteration process. AccSMBO places particular emphasis on the best hyperparameter high probability ranges: modifying $acquisition$ $functions$ in SMBO basing on the  metalearning dataset. We name  our modified $acquisition$ $functions$ as meta-$acquisition$ $functions$, abbr. metaAC function.

\section{Methodology in accSMBO}
AccSMBO uses the following two methods to accelerate the SMBO algorithm frame.
\subsection{Using Gradient-based  multikernel $\mathcal{GP}$ as $\mathcal{M}_L$}
The gradient-based $\mathcal{GP}$ regression  has proven to be an effective $\mathcal{M}_L$ \cite{Wu2017Bayesian}. However, in traditional gradient-based $\mathcal{GP}$ regression,   the unstable character of the local performance curve would mislead the process of building  $\mathcal{M}_L$. What is more, the computational load for traditional gradient-based $\mathcal{GP}$ regression is large.  To deal with the locally unstable problem, we designed a new gradient-based $\mathcal{GP}$. Our gradient-based multikernel $\mathcal{GP}$ regression is different with current multikernel $\mathcal{GP}$ regression and gradient $\mathcal{GP}$ in terms of the initial idea and algorithm\cite{Jie2013A} \cite{Wu2017Bayesian}.

\begin{figure*} 
	\caption{Accurate gradient fitting would mislead regression function}
	\centering
	\subfigure[Regression function with accurate gradient fitting lose function general trend]{
		\begin{minipage}[b]{0.48\textwidth}
			\includegraphics[width=1\textwidth]{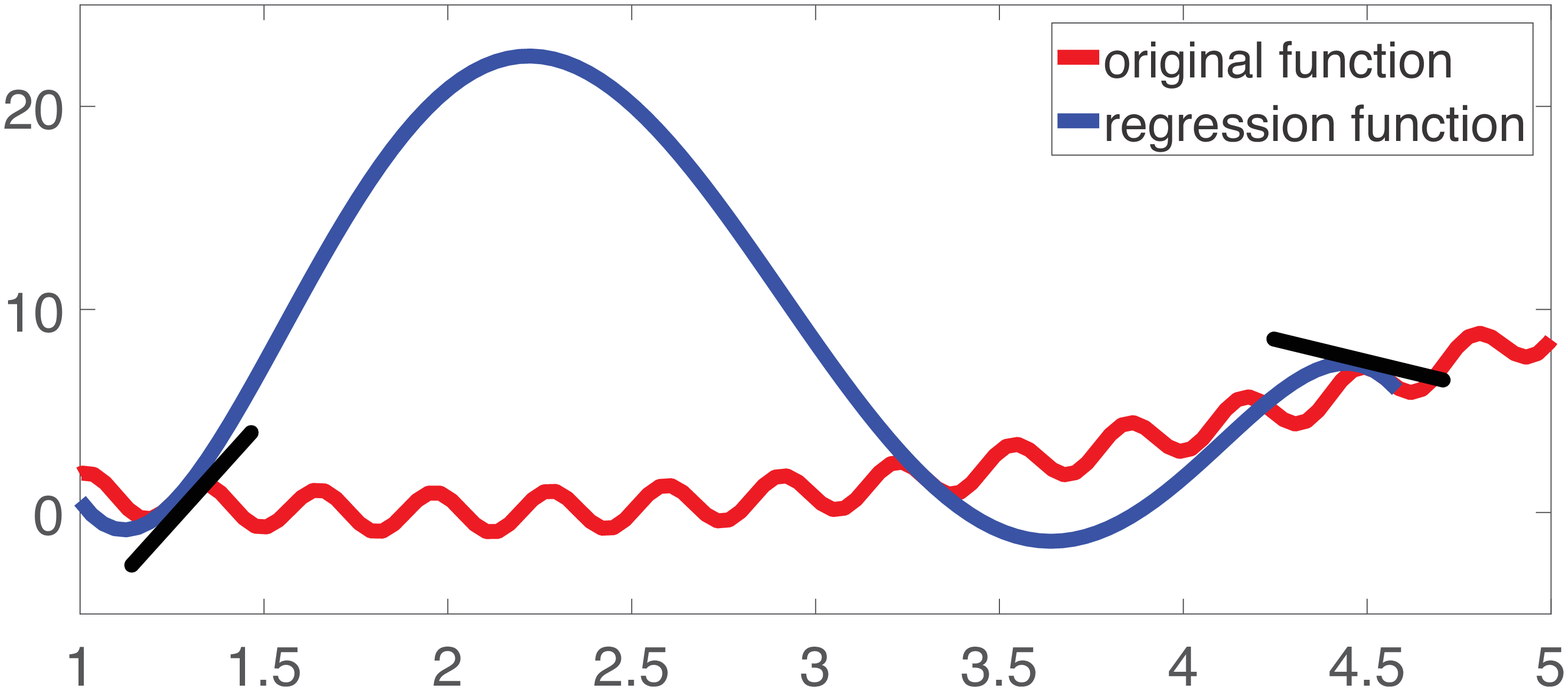}  
			
		\end{minipage}
	}
	\subfigure[Regression function with approximate gradient fitting presents  function general trend]{
		\begin{minipage}[b]{0.48\textwidth}
			\includegraphics[width=1\textwidth]{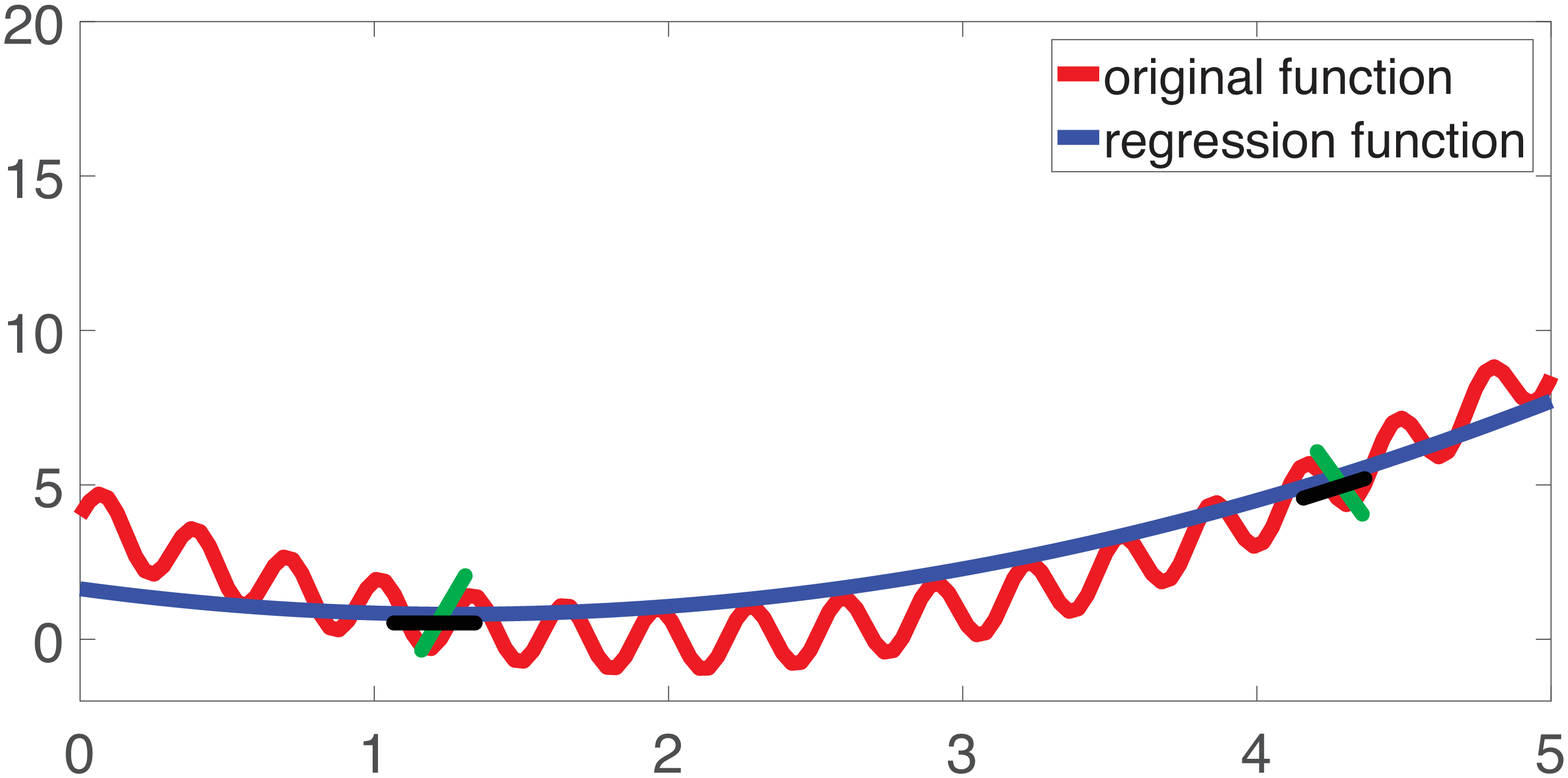} 
			
		\end{minipage}
	}
	\label{tidai}
\end{figure*}

\textbf{Extra notes} 
The dimension of $\lambda$ is d. When the gradient information can be computed, the SMBO history is as follows:

\begin{equation*}
	\mathcal{H} = \{  (\lambda_1,f(\lambda_1),\nabla f(\lambda_1)),...,(\lambda_n,f(\lambda_n),\nabla f(\lambda_n))
	\} 
\end{equation*}

We define  the vector $\nabla \textbf{f}=(\nabla f(\lambda_1), \nabla f(\lambda_2),...,\nabla f(\lambda_n))^T$, whose dimension is (d * n). We also define the matrix $\bm{\lambda} = ( \lambda_1, \lambda_2,...,\lambda_n)$, whose dimension is (d*n), the kernel matrix $K(\bm{\lambda},\bm{\lambda})$, whose dimension is (n*n), and the gradient kernel matrix $\nabla K(\bm{\lambda},\bm{\lambda})$ whose dimension is (n*(d*n)) as follows:

\begin{equation*}
	\nabla K(\bm{\lambda},\bm{\lambda}) =
	\left[
	\begin{matrix}
		\nabla k(\lambda_1,\lambda_1)      &\nabla k(\lambda_1,\lambda_2)       & \cdots & \nabla  k(\lambda_1,\lambda_n)       \\
		\nabla k(\lambda_2,\lambda_1)      & \nabla k(\lambda_2,\lambda_2)       & \cdots &\nabla  k(\lambda_2,\lambda_n)       \\
		\vdots & \vdots & \ddots & \vdots \\
		\nabla k(\lambda_n,\lambda_1)       &\nabla  k(\lambda_n,\lambda_2)       & \cdots & \nabla k(\lambda_n,\lambda_n)       \\
	\end{matrix}
	\right]
\end{equation*}

And $K_{m:n}=(K_m,K_{m+1},K_{m+2},...,K_{n})$ is the abbr. of combination of matrix.

\textbf{Gradient-based multikernel $\mathcal{GP}$ regression}  In this paper, we offer an innovative  gradient-based $\mathcal{GP}$ regression.  The solving process is shown in the algorithm \ref{gradient_algorithm}. 

In this process, we extract $\bm \lambda$, $\nabla \bm f$ and $\bm f$ from $\mathcal{H}$. Based on that information, we want to build the gradient-based multikernel $\mathcal{GP}$ regression whose mean function's gradient is closest to the observations.

The essence of our method, i.e. using the combination kernel on the mean function, is the ordinary  $\mathcal{GP}$ combination, i.e.,

\begin{equation*}
	\mathcal{M}_L = \mathcal{GP}_{combine} = \mathcal{GP}_{k_1} +  \mathcal{GP}_{k_2} +...+ \mathcal{GP}_{k_{d+1}}
\end{equation*}

where $\mathcal{GP}_{k_1}$ denotes the $\mathcal{GP}$ which is regressed by kernel $k_1(\cdot,\cdot)$. The above view shows that observations are produced by the sum of the different $\mathcal{GP}$es, which are regressed by different kernels.

Compared with the traditional $\mathcal{GP}$, which is regressed by only one kernel, the $\mathcal{GP}$ combination has higher degrees of freedom: d+1 variable $\alpha$. Those extra degrees of freedom allow our method to fit the observed information such as the observed gradient and point values.

\textbf{Mean} \textbf{function}  In the traditional $\mathcal{GP}$, the mean function $m(\lambda)$ is a fitting function that uses kernel functions as a basis.

To implement the basic idea presented above, we initially combined different kernels, linear independent kernels, with different coefficients to fit the equations \ref{grad_equ} and \ref{point_equ}. When the number of kernels is equal to the dimension of $\nabla \bm f$, the mean function can reflect all the observed information, including point and gradient observed values.

\begin{align}
	&\bm f=K_1(\bm \lambda,
	\bm \lambda) \bm{\alpha_1} + K_2(\bm \lambda,
	\bm \lambda)  \bm{\alpha_2}  ...K_{d+1}(\bm \lambda,
	\bm \lambda)   \bm{\alpha_{d+1}}\label{point_equ}  \\
	&\nabla \bm f = \nabla K_1(\bm \lambda,
	\bm \lambda) \bm{\alpha_1}  + ... 
	\nabla K_{d+1}(\bm \lambda,
	\bm \lambda)\bm{\alpha_{d+1}} \label{grad_equ}
\end{align}

When the Eq. \ref{grad_equ} and \ref{point_equ} are solved, we can obtain $\bm \alpha_1, \bm \alpha_2,\cdots,\bm \alpha_{d+1}$. The mean function, $m(x)$, of the $\mathcal{GP}$ can be described as $m(x)=\sum_{i=1}^{d+1} K_i(\lambda^*,\bm \lambda) \bm \alpha_i$

\textbf {variance function} The variance function of those $\mathcal{GP}$es reflects the distance of the prediction point $\lambda^*$ and the observed point $\lambda_i$. Thus, for $\mathcal{GP}_{k_i}$,  the variance function is $var_i(\lambda^*) = k_i(\lambda^*,\lambda^*) - K_i(\bm{\lambda},\lambda^*)K_i(\bm{\lambda},\bm{\lambda})^{-1}K_i(\bm{\lambda},\lambda^*)$. Because the sum of $\mathcal{GP}$es is still a $\mathcal{GP}$, $\mathcal{GP}_{combine}$ is $\mathcal{N}(\sum_{i=1}^{d+1}m_i(\lambda^*),\sum_{i=1}^{d+1}var_i(\lambda^*))$

\textbf{Generalization, approximate and reduce the computational load} Although the radial basis function (RBF) kernel satisfies all the requirements listed above (such as polyharmonic spline functions), we still need to address the computational load and generalisation problem. As we mentioned in the section "Main Idea", the performance curve is simple but full of waves. The gradients reflect local information rather than the general trend of the performance curve. Accurate gradient information and fitting introduce a negative influence in $\mathcal{GP}$ regression and increase the computational load. In particular, when the gradient of the sampled point is against the general trend, the negative influence of the regression process would be significant, shown in Figure \ref{tidai}.  Thus, an appropriate approximation is needed before running the algorithm.

\begin{algorithm} 
	\caption{Gradient-based Gaussian process regression in Algorithm \ref{accSMBO}}
	\label{gradient_algorithm}
	\begin{algorithmic}
		\STATE {\bfseries Input:} {kernel $K_1$, $K_2$,..,$K_n$; History $\mathcal{H}$, which contains $\bm \lambda$ and $\bm f$}
		\STATE {\bfseries Output:} {Gaussian process $\mathcal{GP} $, with the mean function $m(x)$ and covariance function $var(x)$ }
		\STATE    Compute the vector $\bm \alpha$ and $\bm \beta$:
		\begin{small}
			\begin{align}
				& ( \nabla K_{2:n}(\bm \lambda,
				\bm \lambda)    -  K_{2:n}(\bm \lambda,
				\bm \lambda) K_1^{-1}(\bm \lambda,
				\bm \lambda)     \nabla K_1(\bm \lambda,
				\bm \lambda)                      )\bm \alpha \notag\\
				&=\nabla \bm f -  K_{2:n}(\bm \lambda,
				\bm \lambda) K_1^{-1}(\bm \lambda,
				\bm \lambda)     \nabla K_1(\bm \lambda,
				\bm \lambda)  \bm f  \label{sovleLE}\\
				& K_1 (\bm \lambda,
				\bm \lambda) \bm \beta  = \bm f - K_{2:n} (\bm \lambda,
				\bm \lambda) \bm \alpha\notag
			\end{align}
		\end{small}
		\STATE    The Eq. \ref{sovleLE} is an overdetermined equation that can be solved via the least squares method.\;

		\STATE 
		$m(\lambda^*) =  K_{2:n}(\lambda^*,\bm \lambda) \bm \alpha +  K_1(\lambda^*,\bm \lambda) \bm \beta$\;
		
		\STATE     $var(\lambda^*)=  \sum_{i=1}^{n} (k_i(\lambda^*,\lambda^*)-K_i(\bm \lambda,\lambda^*)K_i^{-1}(\bm \lambda,\bm \lambda)K_i(\bm \lambda,\lambda^*)) $\;
		
		\STATE    return $\mathcal{N}(m(\lambda^*),var(\lambda^*))$\;

	\end{algorithmic}
\end{algorithm}

\begin{figure*} 
	\caption{The change of AC function to Meta-AC function}
	\centering
	\subfigure[oringinal objective function and AC function]{
		\begin{minipage}[b]{0.45\textwidth}
			\includegraphics[width=1\textwidth]{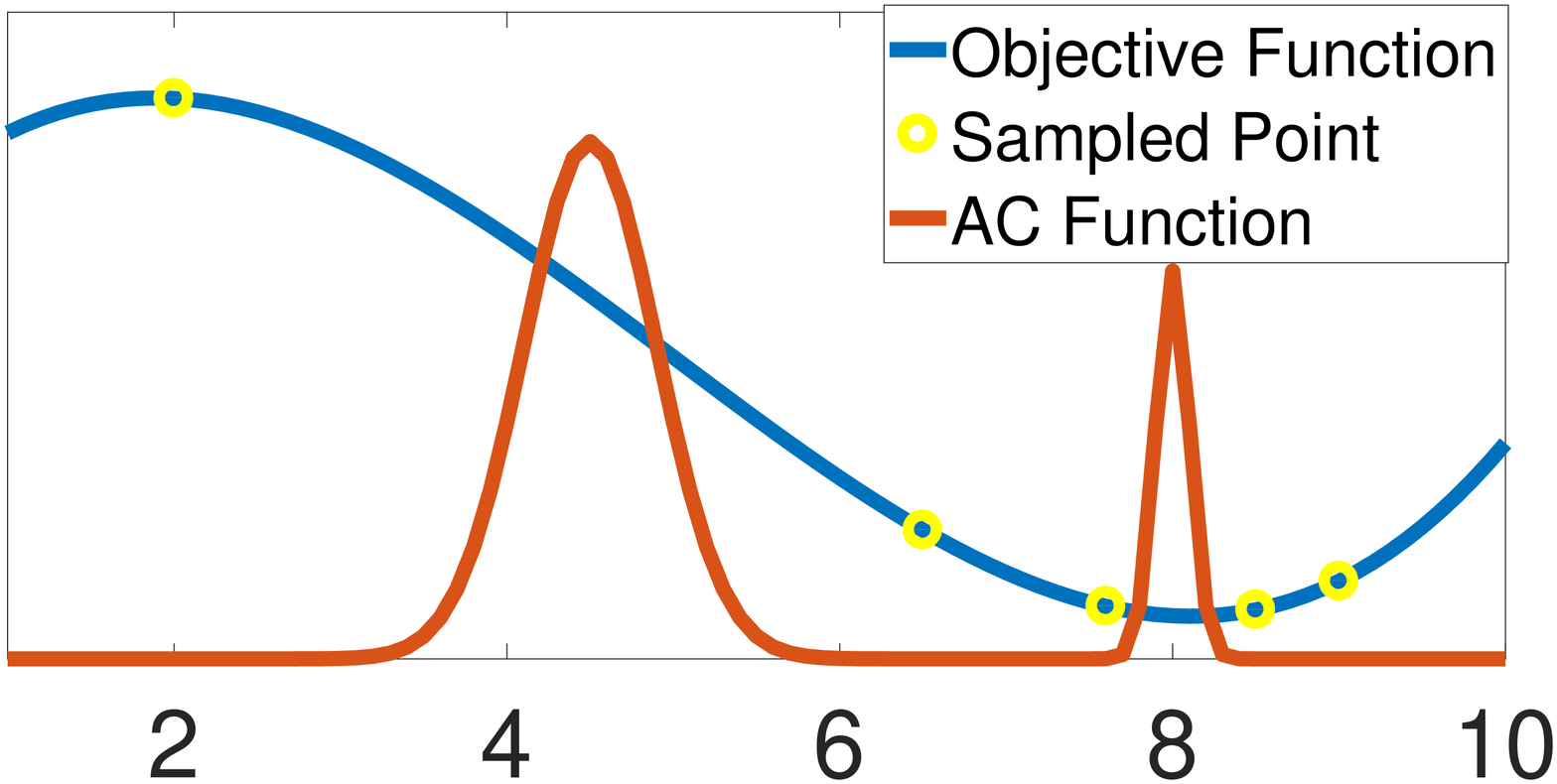}  
		\end{minipage}
	}
	\subfigure[Empirical 
	Probability  Density Function (EPDF) based on Metalearning Dataset]{
		\begin{minipage}[b]{0.45\textwidth}
			\includegraphics[width=1\textwidth]{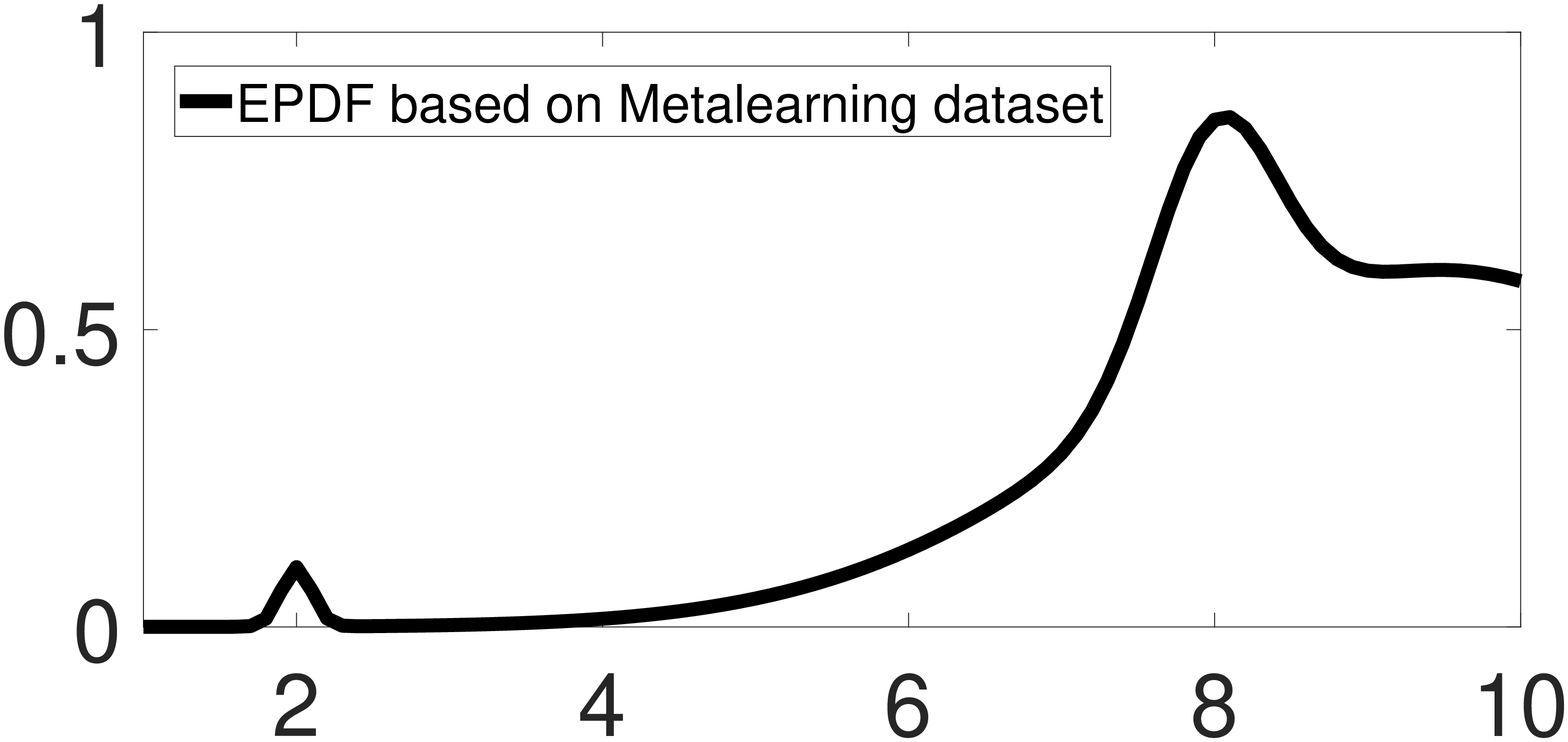}     
		\end{minipage}
	}
	\subfigure[Meta-AC function ]{
		\begin{minipage}[b]{0.45\textwidth}
			\includegraphics[width=1\textwidth]{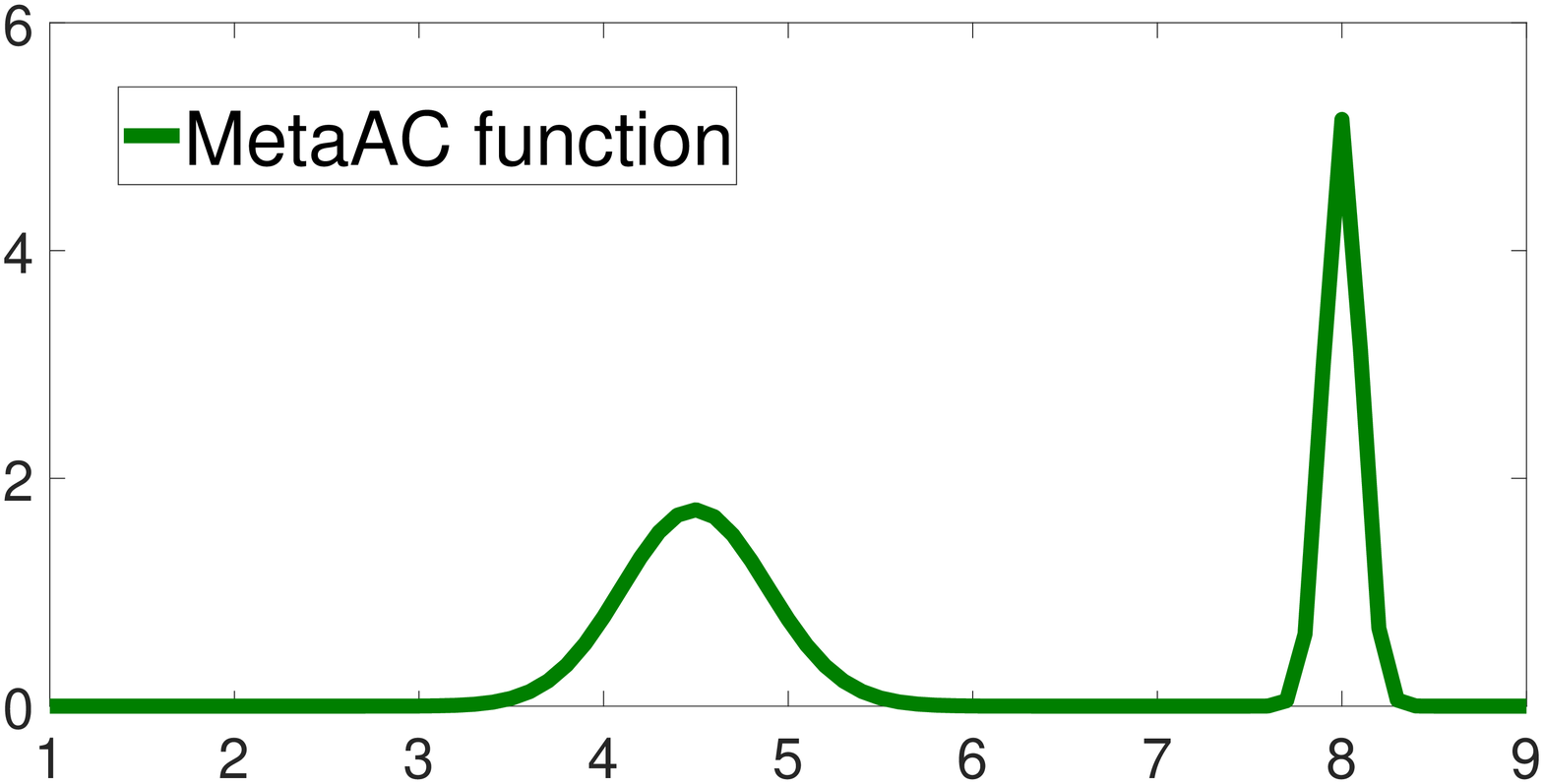}      
		\end{minipage}
	}
	\subfigure[original objective function and Mata-AC function]{
		\begin{minipage}[b]{0.45\textwidth}
			\includegraphics[width=1\textwidth]{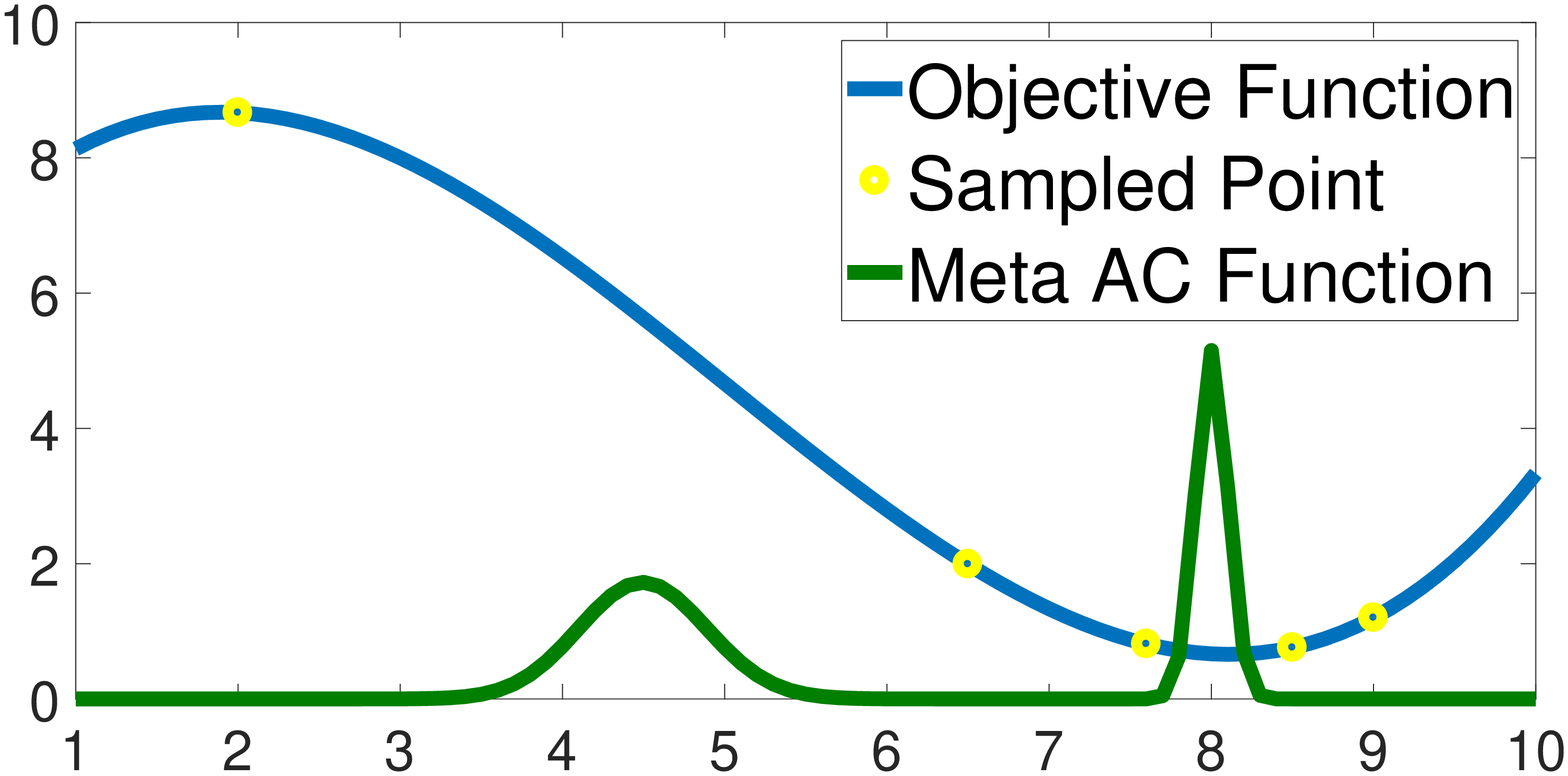} 
			
		\end{minipage}
	}
	\label{meta-process}
\end{figure*}

In our method, we approximate gradient information by reducing the number of kernels. In practice, the more waves, the less number of kernels we should use. Eq. \ref{point_equ} and Eq. \ref{grad_equ} are the difference in status when using the approximation method. For most cases, the performance waves locally would exert more influence on the gradient information, Eq. \ref{grad_equ} instead of points information, Eq. \ref{point_equ}. Thus, we expected that Eq. \ref{point_equ} is accurate and the Eq. \ref{grad_equ} is approximated, i.e. using the least squares method to address the gradient equation (i.e., Eq. \ref{grad_equ}) in the subspace of Eq. \ref{point_equ}.  The Eq. \ref{sovleLE} in algorithm \ref{gradient_algorithm} shows this process. We omit the proof of our process for the limitation of pages and the obviousness of the proof.

Thus, considering the property of the hyperparameter curve and computational load, our gradient-based multikernel $\mathcal{GP}$ regression is a state-of-the-art choice for SMBO.

\subsection{Using Meta-$acquisition$ $functions$ in SMBO process}

Researchers proposed SMBO for the case where the users do not possess any information about the objective function.  However, hyperparameter optimisation is not suited to this case. We often know some prior information about hyperparameter. The metalearning datasets often reflect this prior information. We propose meta-$acquisition$ $function$ to use that information. Meta-$acquisition$ $function$  makes  the $acquisition$ $functions$  high at  the best hyperparameters high probability range

\subsubsection{meta-$acquisition$ $function$}  To make  the $acquisition$ $functions$  high at  the best hyperparameters high probability range, we propose the following method to adjust  $acquisition$ $functions$, and we named our method, i.e. step 2 in algorithm \ref{accSMBO} as meta-$acquisition$ $functions$, abbr. metaAC function.

\textbf{Empirical probability density function }  Before we design metaAC function, it is necessary to fit an empirical probability density function. 

We build frequency  histogram based on the metalearning dataset as our adjustment reference. AccSMBO does not ask we have a complete metalearning dataset which contains every initial hyperparameter values for different situations. Our methods only ask that the metalearning dataset reflects the trend of distribution. To make the following step clearly, we fit above  frequency  histogram into empirical probability density function, abbr. EPDF, $p(x)$. This step is shown in subfigure (b) in Figure \ref{meta-process}.

\textbf{Design metaAC function} To make use of $p(x)$, we expect that accSMBO builds metaAC function follows two principles: 

1). The $p(x)$ encourages SMBO samples more hyperparameter at the best hyperparameter  high probability ranges. As we mentioned above, the distribution of the best hyperparameter is not a uniform distribution. We should explore the best hyperparameter  high probability ranges firstly.

2). With the algorithm processing, the influence of $p(x)$ is decreasing. Overly depending on $p(\lambda)$ would largely break the exploitation and exploration trade-off which would keep SMBO from seeing all scope of objective function: After exploring the best hyperparameter  high probability ranges, we still should sample hyperparameters at other ranges. This requirement  ensures that SMBO gains the whole scope of the objective function and gains the best hyperparameters with the epoch's going to infinite.

Thus, we design following  $metaAC(\lambda,epoch,p(x),ac(\lambda))$ functions:

\begin{align*}
	&metaAC(\lambda,epoch,p(x),ac(\lambda))\\
	&=ac(\lambda)*(rate*p(\lambda) e^{-epoch} +  1-rate*e^{-epoch})
\end{align*}

where $rate$ is the parameter which $ rate\in[0,1] $ and decides the influence of $p(\lambda)$. $rate$ should be larger when the metalearning dataset is complete. 

Above descriptions can be shown in (c) subfigure in Figure \ref{meta-process}. In this case, $rate = 1$.

\textbf{Convergence proof of metaAC} Our method is a modification of original AC function, and with the algorithm process, $metaAC$ degenerates into original AC function. In another word, the convergent character of $metaAC$ is the same as the original AC function with $epoch$'s going to infinite. Thus, the proof of convergent would be the same as the AC function.

\textbf{Acceleration for SMBO} In practice,  we are more likely to meet the cases where the best hyperparameter value is in the high probability ranges.  As we can see from (a) subfigure and (d) subfigure in Figure \ref{meta-process}, after the adjustment by $p(x)$, SMBO pays more attention to the high probability ranges. Thus, metaAC function would accelerate the SMBO process.

\subsubsection{The limitation of current metalearning} Metalearning has been combined into SMBO in recent years. However, current metalearning technology has two shortages. 

(1) Current automl algorithms only use metalearning in offering an optimal/sub-optimal initial hyperparameter.  Metalearning is unable to exert influence on the process of optimisation. 

(2) The metalearning dataset must be large. When the metalearning dataset is too small, metalearning fails to accelerate hyperparameter optimisation: 1) A small metalearning dataset cannot offer the best hyperparameter initial values for some cases. 2) In the SMBO process, to trade off the exploration and exploitation problem, $\mathcal{M}_L$ and traditional $acquisition$ $functions$  encourage SMBO to explore unsampled ranges, which may be the best hyperparameters low probability range. Thus, traditional  $acquisition$ $functions$ are high at the best hyperparameters low probability range. This process would be shown in (a) subfigure in Figure \ref{meta-process}: For unsampled range [2,6], the AC function is high, because this range is unexplored.

\textbf{Overcome current metalearning shortage} Our metaAC function overcomes the shortages of current metalearning technology: 

(1) MetaAC function is part of the optimisation process.

(2) In metaAC function, the metalearning dataset can be small. MetaAC only requires that the metalearning datasets reflect the trend of the best hyperparameters. The missing of some records cannot change the characters of the whole hyperparameter trend.

\section{Experiment }

We use the HOAG experimental framework \cite{Pedregosa2016Hyperparameter} in the experiments.

\subsection{Experiment Settings}
\textbf{Dataset} First, we choose a small dataset  (pc4 in openML), a medium-sized dataset (rcv1) and a large dataset (real-sim).  Information concerning these datasets is listed in Table \ref{dataset}. 

We assume that the gradient in the dataset is valid. In all cases, the dataset is randomly split into two parts: a training set containing 70\% of the dataset samples and a valid set containing 30\% of the dataset samples.

\begin{small}
	\begin{table}[htbp]
		
		\caption{Dataset information}
		\label{dataset}
		\begin{tabular}{|c|c|c|c|c|}
			\hline
			dataset & \#features & \#size & feature range & sparsity \\
			\hline
			real-sim & 20,958& 72,309&(0,1)&sparse\\
			\hline
			rcv1&47,236 &20,242 &(0,1)&sparse\\
			\hline
			pc4&38&1,458&(0,10,000)&dense \\
			\hline    
		\end{tabular}
	\end{table}
\end{small}

\textbf{Problem} In our experiment, we will solve the problem of determining the regularisation parameter in the L2 norm logistic regression model because it is easy to acquire the hyperparameter gradient of the L2 norm\cite{Do2007Efficient}. In this case, the loss function of the inner optimisation problem, i.e., $h(\lambda)$, is the regularised logistic loss function. For classification, the outer cost function is the logistic loss function (i.e., Eq. \ref{experiment}). The logistic loss function overcomes the problem that zero-one loss is a non-smooth loss function\cite{Do2007Efficient}:

\begin{small}
	\begin{align}
		& \mathop{argmin}_{\lambda \in \Lambda} f(\lambda) = \sum_{i=1}^{m}\Phi(x_iy^T_iA(\lambda)) \notag \\
		&s.t. A(\lambda) \in \mathop{argmin}_{\bm{model} \in \mathbb{R}}\sum_{i=1}^{m} \Phi(x_iy^T_i\bm{model}) + \lambda \left\| \bm{model}    \right\| ^2
		\label{experiment}
	\end{align}
\end{small}

where $\Phi$ is the logistic loss, i.e., $\Phi(t) = log(1+e^{-t})$. The solver used for the inner optimization problem of the
logistic regression problem is stochastic gradient descent\cite{Duchi2016Intro}. In our problem, we set the search range to $\lambda \in [0,1]$.

\textbf{Kernel} In this paper, the hyperparameter dimension, i.e., $\lambda$, is one; thus, we can choose either of two kernels for our gradient $\mathcal{GP}$ method: the Gaussian radial basis function $k_2(x_1,x_2) = exp(-\left\| x_1 - x_2 \right\|^2 )$ and the cubic radial basis function $k_1(x_1,x_2) =\left\| x_1 - x_2 \right\|^3$ \cite{Regis2005Constrained}. In our experiment, the inverse $K_2$ matrix calculation is the key aspect of the algorithm. However, the accuracy of $K^{-1}_2$ based on the cubic radial basis function is reduced when $\lambda \in [0,1]$. Thus, we choose the Gaussian radial basis function, $k_2(\cdot,\cdot)$.

\textbf{Metalearning Dataset} The only open metalearning dataset we can find is the metalearning dataset in auto-sklearn\cite{thornton2013auto-weka:}. This metalearning dataset is built by the OpenML dataset. To make our experiments persuasive, we randomly delete 20\% contents in this dataset, because our experiment dataset is also from OpenML. We named our dataset as the half-metalearning dataset. The half-metalearning dataset cannot be used in offering the best initial hyperparameters for it misses the best  hyperparameters for some situations. However, half-metalearning dataset still can be used in our $metaAC$ function for it does not lose the trend character.

\textbf{Comparison with other hyperparameter optimisation methods} In this section, we compare accSMBO against the following five existing hyperparameter optimisation methods. To make the convergence process clear, we turn off metalearning technology in choosing initial value step and all initial value $\lambda$ is set as 1.

$SMAC$, a state-of-the-art method, performs SMBO using a random forest.  SMAC is the core algorithm in autosklearn. The initial hyperparameter for this method is $\lambda = 1$. We select four challengers in each epoch using the intensify process to choose the single best challenger (this is the default setting for autosklearn).  The acquisition function used in these experiments is the expected improvement function.  As we mentioned in the background section, SMAC is the main benchmark for our AccSMBO.

$AccSMBO$   In this paper, we modified the autosklearn framework.  In the setting of our experiment, we set $rate$ in $metaAC$ function as 1.  To make $p(x)$ accurate, we build $p(x)$ on the metalearning data after deciding object, task and the feature of the dataset. For example, in this experiment cases, we build $p(x)$ based on the metalearning data for the case where 1). The objective function is logloss, 2) the task is binary classification and 3). The dataset is sparse dataset.     

We select four challengers in each epoch and use the intensify process to choose the single best challenger (this is also the default setting for autosklearn).

$Grid$ $Search$, common method. We adopt the method we presented in this paper, with an exponentially decreasing tolerance sequence. In the interval $[0,1]$, we sample 20 $\lambda$ hyperparameter uniformly; then, we compute the performance of $f(\lambda)$ serially from $\lambda = 1$ to $\lambda = 0$.

$Random$ $Search$, common method. This random search method samples the hyperparameters from a predefined distribution. We choose the samples from a uniform distribution in the interval $[0,1]$. To ensure that all methods have similar initial values, we compute the performance in the first epoch with $\lambda = 1$.

$HOAG$ is a state-of-the-art method which uses gradients to find the minimum  of the objective function, similar to gradient descent. In each epoch, the hyperparameter is adjusted toward the gradient direction. Here, we use the modified HOAG framework from the work\cite{Pedregosa2016Hyperparameter}. In the initial hyperparameter for this method, $\lambda = 1$. To achieve the fastest convergence speed in each epoch, we set $\epsilon_k = 10^{-12}$. To measure the HOAG's convergence speed, we learned the performance curve before the HOAG experiments. Based on the Lipschitz constant in $[0,1]$, the step length in HOAG is fixed to $10^{-3}$ in all the experiments.

Above benchmarks would show that accSMBO is better than the widely used method (random search, grid search), gradient-based method (HOAG) and other SMBO algorithm( SMAC).

\subsection{Experimental results and analyses}

Figure \ref{pc4exp} shows the experimental results on the pc4 dataset. Our methods achieve the fastest convergence speed and find the best hyperparameter, while the output from SMAC is suboptimal. Our method achieves convergence 400\% faster in epoch than SMAC.  When the dataset scale is small, the performance curve characteristic is unstable and has a multimodal function. HOAG also results in poor performance. The output of HOAG does not appear to have converged. The performance curve on the pc4 dataset is unstable and non-smooth, and the gradient information does not indicate the trend of the curve. Because HOAG uses an accurate gradient as its optimisation direction, it fails to find the best hyperparameter in this case. Bayesian optimisation shows its advantages here because it ignores the "noise" in the curve. Our method finds the curve function tendency and approximates the gradient more quickly. 

\begin{small}
	\begin{figure}[htbp]
		\centering
		\caption{Algorithm performances on the Pc4 dataset}
		\includegraphics[width=0.47\textwidth]{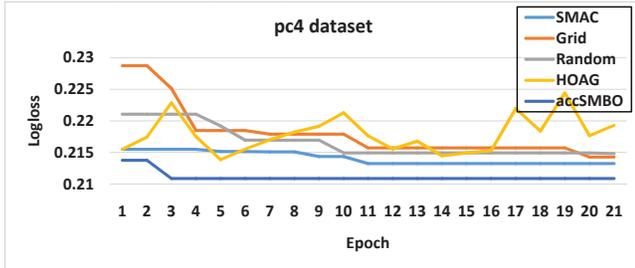} 
		\label{pc4exp}
	\end{figure}
\end{small}

\begin{figure}[htbp]
	\centering
	\caption{Algorithm performances on the rcv1 dataset}
	\includegraphics[width=0.47\textwidth]{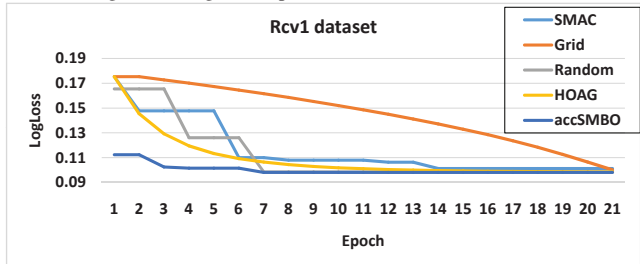} 
	\label{rcv1exp}
\end{figure}

Figure \ref{rcv1exp} shows the experimental results on the rcv1 dataset. Again, our methods achieve the fastest convergence speed and find the best hyperparameter. In this case, the performance curve characteristics are relatively stable and close to a unimodal function. The output from HOAG is suboptimal. Our method achieves convergence 140\% faster in epoch than HOAG and 200\% faster in epoch than SMAC.  SMAC's result is almost the same as a random search because SMAC requires more sample points to build the information for the entire curve.

Figure \ref{real-simexp} shows the experimental results on the real-sim dataset. Here, HOAG and our methods achieve the fastest convergence speed and find the best hyperparameter. accSMBO achieves convergence 300\% faster in epoch than SMAC.    
\begin{small}
	\begin{figure}[htbp]
		\centering
		\caption{Algorithm performances on the real-sim dataset}
		\includegraphics[width=0.47\textwidth]{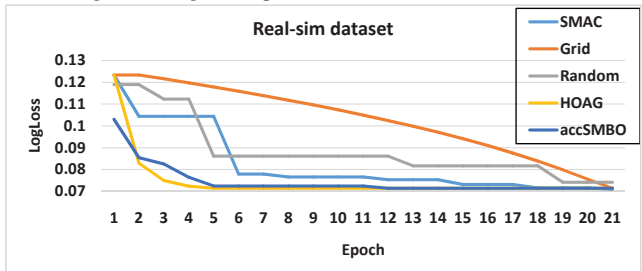} 
		\label{real-simexp}
	\end{figure}
\end{small}



\section{Conclusion }

In this paper, we use two methods to accelerate SMBO. 1) We use a gradient-based multikernel $\mathcal{GP}$ to build $\mathcal{M}_L$, 2)  We design and use the metaAC function. In the experiments, our methods achieved state-of-the-art performances, converging 140\% to 300\% faster than SMAC algorithm on the pc4, real-sim and rcv1 datasets. In many cases, our method outperformed the previous best hyperparameter optimisation approach.

\bibliographystyle{IEEEtran}
\bibliography{IEEEabrv,mybibfile_graybox}
\end{document}